\documentclass[times, review, 10pt]{elsarticle}

\usepackage{amsmath, bm, amssymb, amsthm}

\newtheorem{hypothesis}{Hypothesis}

\usepackage[caption=false,font=footnotesize]{subfig}
\usepackage{float}
\usepackage{threeparttable} 
\usepackage{tabularray}
\usepackage{csquotes}
\usepackage{textcomp}
\usepackage{url}

\usepackage[ruled, linesnumbered, noend]{algorithm2e}
\usepackage{letltxmacro}        
\usepackage{xparse}
\SetAlCapNameFnt{\small}
\SetAlCapFnt{\small\normalfont}
\SetCommentSty{footnotesize}{}
\SetArgSty{textrm}

\ExplSyntaxOn
\let\originalSetKwInput\SetKwInput

\RenewDocumentCommand{\SetKwInput}{m m}
{
	\expandafter\NewDocumentCommand\csname #1\endcsname{O{#2} m}
	{
		\originalSetKwInput{@temp@kin}{##1}
		\csname @temp@kin\endcsname{##2}
	}
}
\ExplSyntaxOff

\SetKwInput{MyKwIn}{Input}
\SetKwInput{MyKwOut}{Output}

\makeatletter
\@ifpackagelater{algorithm2e}{2017/07/19}{}{
	\renewcommand{\algocf@Vline}[1]{
		\strut\par\nointerlineskip
		\algocf@push{\skiprule}
		\hbox{\kern-.4pt\vrule
			\vtop{\algocf@push{\skiptext}
				\vtop{\algocf@addskiptotal #1}\Hlne}}\vskip\skiphlne
		\algocf@pop{\skiprule}
		\nointerlineskip}
	\renewcommand{\algocf@Vsline}[1]{
		\strut\par\nointerlineskip
		\algocf@bblockcode%
		\algocf@push{\skiprule}
		\hbox{\kern-.4pt\vrule
			\vtop{\algocf@push{\skiptext}
				\vtop{\algocf@addskiptotal #1}}}
		\algocf@pop{\skiprule}
		\algocf@eblockcode%
	}
}
\makeatother

\LetLtxMacro\oldalgorithm\algorithm          
\LetLtxMacro\endoldalgorithm\endalgorithm    

\renewenvironment{algorithm}[1][]{%
	\oldalgorithm[#1]              
	\linespread{0.93}\selectfont   
}{%
	\endoldalgorithm               
}

\usepackage[utf8]{inputenc}     
\usepackage[T1]{fontenc}        
\usepackage{listings}           

\usepackage{courier}                

\usepackage{xcolor}       
\definecolor{codeblue}{rgb}{0,0,0.8}
\definecolor{codegreen}{rgb}{0,0.5,0}
\definecolor{codegray}{rgb}{0.5,0.5,0.5}

\begin{document}

\begin{frontmatter}
	
\title{AdamNX: An Adam improvement algorithm based on a novel exponential decay mechanism for the second-order moment estimate}  
	
\author[1]{Meng Zhu}
\ead{zhumeng@jxufe.edu.cn}
\author[1]{Quan Xiao}
\ead{xiaoquan@foxmail.com}
\author[2,3,4]{Weidong Min\corref{corresponding}}
\ead{minweidong@ncu.edu.cn}
	
\affiliation[1]{organization={School of Information Management and Mathematics, Jiangxi University of Finance and Economics},
	city={Nanchang},
	citysep={},
	postcode={330032}, 
	country={China}}
\affiliation[2]{organization={School of Mathematics and Computer Science, Nanchang University},
	city={Nanchang},
	citysep={},
	postcode={330031}, 
	country={China}}
\affiliation[3]{organization={Institute of Metaverse, Nanchang University},
	city={Nanchang},
	citysep={},
	postcode={330031}, 
	country={China}}
\affiliation[4]{organization={Jiangxi Provincial Key Laboratory of Virtual Reality},
	city={Nanchang},
	citysep={},
	postcode={330031}, 
	country={China}}
	
\cortext[corresponding]{Corresponding author}
	
\begin{abstract}	
	This paper studies the exponential decay mechanism of the second-moment estimate in Adam. We propose AdamNX and a time-varying second-moment decay rate that gradually weakens the correction applied to the update scale. Under the assumptions used in our analysis, this mechanism makes the updates approach momentum-SGD-like behavior during the training plateau phase. We report results on the image-classification, object-detection, and semantic-segmentation tasks, configurations, and comparators included in this paper. These results do not establish multi-seed statistical effects, flatness, or generalization beyond the reported tasks. Our code is open-sourced at \url{https://github.com/mengzhu0308/AdamNX}.
\end{abstract}
	
	
\end{frontmatter}

\section{Introduction}\label{sec1}

This paper focuses on Adam's second-moment estimate and its effect on update scaling, which defines the scope of the AdamNX design.

Table~\ref{tbl-optimization-algorithm-iteration} chronologically outlines the milestones of optimization algorithms in deep learning. In 1986, Rumelhart~\emph{et~al.}~\cite{1986-Rumelhart-GD-SGD} first proposed an efficient algorithm to calculate the exact gradient of all weights in a multilayer network on the full dataset in one go. For large-scale datasets, the cost of iterating once with gradient descent (GD) is too high, \emph{i.e.}, it requires too much computational resources. To address this issue, they also proposed the SGD algorithm at the same time, which randomly draws a small batch of data from the training dataset to calculate the gradient and update the model parameters. However, there are three problems for SGD: (1) Lack of \enquote{momentum}\textemdash slow convergence. Each step is taken only in the direction of the current gradient, which easily causes oscillation back and forth on the narrow \enquote{valley} shaped surface, resulting in more convergence steps. (2) Unequal learning step sizes for different parameters. When encountering sparse features or scenarios with large gradient scale differences, the direction with large gradient oscillates violently, while the direction with small gradient is almost not updated. (3) Sensitive to learning rates. A slightly larger learning rate can easily cause divergence, while a slightly smaller one can easily cause stagnation. To address problems 1 and 3 of SGD, Sutskever~\emph{et~al.}~\cite{2013-Sutskever-SGDM} introduced momentum into SGD, using \enquote{inertia} to suppress oscillations and accelerate traversal through flat areas, thus making it more robust to learning rates. To address problems 2 and 3 of SGD, Tieleman~\emph{et~al.}~\cite{2012-Tieleman-RMSProp} proposed adaptive step sizes to combat sparse features or scenarios with large gradient scale differences, thus making it more robust to learning rates. However, both momentum SGD and RMSProp algorithms only solved partial problems in SGD, but did not simultaneously obtain the advantages of both. Therefore, Adam~\cite{2015-Kingma-Adam} combined momentum and adaptive learning step size, and made it more robust to learning rates through unbiased estimate, \emph{i.e.}, bias correction. To address the problems of Adam having to cache two additional sets of parameters and high computational complexity, Chen~\emph{et~al.}~\cite{2023-Chen-Lion} proposed the EvoLved Sign Momentum (Lion) algorithm. Compared with Adam, Lion has lower computational complexity and is more memory-efficient. Recently, Liu~\emph{et~al.}~\cite{2024-Liu-Sophia} proposed the Sophia algorithm, which introduces a lightweight Hessian diagonal estimate as a preconditioner and combines a clipping mechanism to control the update step size, making the pretraining speed of language models twice as fast as Adam and reducing the cost by half. Jordan~\emph{et~al.}~\cite{2024-Jordan-Muon} proposed the MomentUm Orthogonalized by Newton-schulz (Muon) algorithm, which is specially customized for matrix parameters.

\begin{table}[H]
	\scriptsize
	\caption{Key nodes and iterations of optimization algorithms in deep learning}
	\label{tbl-optimization-algorithm-iteration}
	\begin{tblr}{
			width=\linewidth, 
			colspec={X[l,0.2] X[l,0.3] X[j] X[j]}, 	
			row{1}={font=\bfseries},             
			hline{1,Z}={1pt},                    
			hline{2}={0.5pt},                    
			columns={valign=m},                  
			rowsep=0.5pt                         
		}
		Year & denoteative algorithms & Core innovations & Main limitations 
		\\
		1986 & GD~\cite{1986-Rumelhart-GD-SGD} 
		& Full-data gradient, theoretically stable 
		& The cost of iterating once on large-scale datasets is too high 
		\\
		1986 & SGD~\cite{1986-Rumelhart-GD-SGD} 
		& Random batch samples for approximation, computationally efficient 
		& (1) Lack of \enquote{momentum}\textemdash slow convergence.
		(2) Unequal learning step sizes for different parameters.
		(3) Sensitive to the learning rate
		\\
		2013 & momentum SGD~\cite{2013-Sutskever-SGDM} 
		& Momentum accumulation, suppression of oscillations, and acceleration of convergence 
		& No adaptive step size
		\\
		2012 & RMSProp~\cite{2012-Tieleman-RMSProp}
		& First-order moment estimate, adaptive learning step size 
		& Lack of directional inertia 
		\\
		2015 & Adam~\cite{2015-Kingma-Adam} 
		& Unbiased estimate of both moments 
		& Generalization disadvantage, additional storage required for first-order and second-order moments 
		\\
		2017 & AdamW~\cite{2017-Loshchilov-AdamW} 
		& Decoupling of weight decay and target updates 
		& Better generalization than Adam, but still requires additional storage for first-order and second-order moments
		\\
		2023 & Lion~\cite{2023-Chen-Lion} 
		& Sign-based Momentum 
		& Not as effective as Adam in small batches 
		\\
		2024 & Sophia~\cite{2024-Liu-Sophia} 
		& Introduction of lightweight diagonal Hessian estimate, combined with a clipping mechanism to control the update step size 
		& (1) Information loss due to diagonal approximation.
		(2) Hard trade-off between estimate frequency and timeliness 
		\\
		2024 & Muon~\cite{2024-Jordan-Muon} 
		& Updates for matrix parameters 
		& Higher computational cost than Adam, increased parallel communication cost 
		\\
	\end{tblr}
\end{table}

Adam maintains first- and second-moment estimates and uses the latter to scale updates. Prior work has discussed different late-stage update characteristics of adaptive and momentum methods. We propose AdamNX (see Algorithm~\ref{algo-adamnx}), which gradually weakens this correction through a time-varying second-moment decay rate. Under the assumptions of our analysis, its plateau-phase updates approach momentum-SGD-like behavior. We do not present this mechanism as a proof of flatness, generalization, or general stability.

The remainder of this paper is organized as follows. Section~\ref{sec2} reviews the existing research on optimization algorithms in deep learning and introduces the motivation of this study. Section~\ref{sec3} provides a detailed analysis of the problem and proposes the corresponding solution. Sections~\ref{sec4}, \ref{sec5}, and \ref{sec6} jointly verify the superiority of AdamNX through experiments. The final section summarizes the content of the entire paper.

\section{Related work}\label{sec2}

\subsection{Optimization algorithms in deep learning}\label{sec2.1}

In 1986, Rumelhart~\emph{et al.}~\cite{1986-Rumelhart-GD-SGD} first proposed the backpropagation algorithm that can calculate the precise gradient of a multilayer network on the entire training set in one go. Although this algorithm is theoretically rigorous, when facing large-scale datasets, the computational and storage costs of a single iteration become a bottleneck. To alleviate this problem, Rumelhart~\emph{et al.}~\cite{1986-Rumelhart-GD-SGD} also proposed to use mini-batch gradient approximation to replace the full gradient in SGD. The core idea is to randomly select a small batch of samples to estimate the gradient in each iteration, thereby significantly reducing the cost of a single parameter update. However, the convergence behavior of SGD in practice is not ideal, mainly manifested in three aspects of defects: (1) Lack of momentum. The update direction completely depends on the current gradient, resulting in frequent oscillations on narrow or high-curvature error surfaces and slow convergence. (2) Inconsistent step size. When the gradient scales of different parameters vary greatly, a uniform learning rate cannot take care of both sparse and dense features at the same time, resulting in some parameters being stagnant in updates and others vibrating violently. (3) Sensitive to learning rate. A slightly larger learning rate can easily cause the objective function to diverge, while a smaller one falls into inefficient updates.

To address defects 1 and 3 of SGD, Sutskever~\emph{et al.}~\cite{2013-Sutskever-SGDM} introduced an exponentially weighted momentum term into SGD, integrating historical gradient information into the current direction to suppress oscillations and accelerate traversal of flat regions. Meanwhile, the robustness to the learning rate is also improved. Gitman~\emph{et al.}~\cite{2019-Gitman-UnderstandingSGDM} used stochastic differential equations to characterize the effects of different momentum values on escaping saddle points and convergence speed, and provided a practical range for momentum selection. Ramezani-Kebrya~\emph{et al.}~\cite{2024-Ramezani-Kebrya-GeneralizationSGDM} provided upper bounds on the stability and generalization error of momentum SGD, theoretically explaining why momentum not only accelerates training but also reduces the risk of overfitting. Zhao~\emph{et al.}~\cite{2024-Zhao-SNGDM} proposed the \enquote{normalization + momentum} combination, proving that it can maintain an $\mathcal{O}(1 / T)$ convergence efficiency under anisotropic noise, effectively alleviating the directional oscillations of SGD.

To address defects 2 and 3 of SGD, Duchi~\emph{et al.}~\cite{2011-Duchi-AdaGrad} proposed the AdaGrad algorithm, which achieves per-parameter adaptive learning step size by accumulating the sum of squares of historical gradients, effectively improving the convergence efficiency in sparse data scenarios. However, the accumulation of the sum of squares of historical gradients as the denominator in AdaGrad causes the learning step size to monotonically decrease to close to zero during the training process, and updates almost stop in the later stages. To address this issue, Tieleman~\emph{et al.}~\cite{2012-Tieleman-RMSProp} proposed RMSProp, which replaces the global accumulation with the exponential moving average of the gradient squares to avoid the learning step size from dropping to zero.

Although momentum SGD and RMSProp each improved some local problems of SGD, their advantages were not obtained at the same time. Kingma~\emph{et al.}~\cite{2015-Kingma-Adam} proposed Adam, which integrates the momentum mechanism and adaptive step size strategy, and further introduces bias correction to offset the systematic bias of the exponentially weighted average, thereby further improving the robustness to the learning rate. Reddi~\emph{et al.}~\cite{2018-Reddi-AMSGrad} proposed the AMSGrad algorithm, which replaces the second-order momentum sliding average in Adam with the maximum value of historical gradients to avoid converging to local extreme solutions with poor generalization. Loshchilov~\emph{et al.}~\cite{2017-Loshchilov-AdamW} proposed AdamW, which decouples weight decay from the target update of the optimization algorithm, making it an independent regularization term from the target update, improving generalization ability, and making hyperparameter tuning more robust. Shazeer~\emph{et al.}~\cite{2018-Shazeer-Adafactor} proposed the Adafactor algorithm, which achieves sublinear memory consumption through low-rank decomposition of the second-order moment, effectively reducing the training memory requirements and is suitable for large model training. Liu~\emph{et al.}~\cite{2024-Liu-Sophia} proposed Sophia, which introduces lightweight diagonal Hessian estimation into large-scale language model pretraining, achieving faster convergence and lower computational cost than Adam. Sophia uses diagonal Hessian approximation instead of the complete Hessian matrix, which reduces the computational cost but sacrifices the curvature correlation information between parameters. The Hessian is estimated every $k$ steps, which keeps the average computational cost within $5\%$ of the gradient calculation, but may lead to dynamic response lag. The estimation variance may be amplified in the early training stage or in small-batch scenarios, affecting stability.

AdamNX differs structurally from SWATS's discrete switch~\cite{2017-Keskar-SwitchingAdamSGD} and AdaI's dynamical modification~\cite{2022-Xie-AdaptiveInertia}: it uses a continuous time-varying second-moment coefficient within an Adam-style update. We retain this structural comparison but do not report experiments against SWATS or AdaI.

In addition, Adam and AdamW need to store the first-order moment estimate and the second-order moment estimate for each parameter, which have high memory occupation and computational complexity. To address the high memory and high computation defects, Chen~\emph{et al.}~\cite{2023-Chen-Lion} proposed Lion, which effectively reduces memory demand and computational volume through symbolic momentum, and can match or even outperform Adam in most tasks. However, when the batch size is small (less than 64), Lion is not as good as Adam. Recently, Jordan~\emph{et al.}~\cite{2024-Jordan-Muon} proposed Muon for matrix parameters, which uses Newton-Schulz iteration to approximately orthogonalize the momentum update matrix, significantly improving the training efficiency of matrix parameters. However, when the matrix is split and distributed across multiple devices, Muon must first aggregate the gradients of each slice (all-reduce) and then calculate the update amount uniformly, which cannot be updated independently in parallel on each device, thus introducing additional communication overhead.

In summary, from full gradient to SGD, and then to momentum, adaptive step sizes, bias correction, and specialized optimizers for structural features, the evolution of optimization algorithms in deep learning has always revolved around \enquote{improving convergence efficiency\textemdash reducing computational costs\textemdash enhancing hyperparameter robustness}.

\subsection{Research motivation}\label{sec2.2}

We consider a continuous second-moment decay schedule. It retains stronger adaptive correction early in training and weakens it later. Under the stated assumptions, it illustrates an Adam-style update approaching momentum-SGD-like behavior. It does not establish algorithmic degeneration or gains in flatness or generalization.

\section{Proposed algorithm}\label{sec3}

This section analyzes the effect of $\beta_{2}$ on the variance of Adam's second-moment estimate under explicit assumptions, then introduces a time-varying second-moment decay rate and AdamNX. The analysis offers a mechanism-oriented explanation only; it does not provide a convergence rate, regret bound, or general stability guarantee. We then compare the second-moment coefficients of AdamNX and Adam.

\subsection{Impact of $\beta_2$ on the variance of the second-order moment estimation in Adam}\label{sec3.1}

Let: 
\begin{equation}
	\bm{g}_t = \nabla f(\bm{\theta}_{t-1}) + \bm{\xi}_t
	\label{eq-sgd-gd-noise}
\end{equation}
where $\nabla f(\bm{\theta}_{t-1})$ denotes the full-batch gradient, \emph{i.e.}, the gradient of $\bm{\theta}_{t-1}$ computed using all samples $\bm{\mathcal{D}}$, $\bm{g}_t$ denotes the stochastic gradient, \emph{i.e.}, the gradient of  $\bm{\theta}_{t-1}$ computed by randomly sampling a subset $\bm{\mathcal{R}} \subset \bm{\mathcal{D}}$ from $\bm{\mathcal{D}}$, and $\bm{\xi}_t$ denotes the noise introduced by random sampling. Thus, $\bm{g}_t$ is an approximation to $\nabla f(\bm{\theta}_{t-1})$. To close the following derivation, we use idealized isotropy, independence, and i.i.d. assumptions. In practical mini-batch training, noise may depend on the parameters, data augmentation, and time, so these assumptions are usually approximations. The following results are used as a heuristic analysis under Hypotheses~\ref{hypothesis-iid}, \ref{hypothesis-noise-norm}, and \ref{hypothesis-grad-noise-independent}:
\begin{hypothesis}\label{hypothesis-iid}
	The full-batch gradient vector $\nabla f(\bm{\theta}_{t-1})$ and the noise vector $\bm{\xi}_t$ have independent and identically distributed (i.i.d.) components across dimensions.
	\footnote{This assumption was implicitly used in the derivations of arXiv v1 but was omitted from the formal statement of assumptions.}
\end{hypothesis}
\begin{hypothesis}\label{hypothesis-noise-norm}
	$\bm{\xi}_t \sim \mathcal{N}(\bm{0}, \sigma^2 \bm{1})$.
\end{hypothesis}
\begin{hypothesis}\label{hypothesis-grad-noise-independent}
	The full-batch gradient $\nabla f(\bm{\theta}_{t-1})$ and the stochastic noise $\bm{\xi}_t$ are mutually independent.
\end{hypothesis}
Based on Hypothesis~\ref{hypothesis-noise-norm}, Equation~\eqref{eq-sgd-gd-noise} can be written as:
\begin{equation}
	\bm{g}_t = \nabla f(\bm{\theta}_{t-1}) + \sigma \bm{\xi}
	\label{eq-sgd-gd-noise-S2}
\end{equation}
where $\bm{\xi}$ denotes the standard normal distribution.

For Adam, the second moment estimate is given by Equation~\eqref{eq-som-estimate}:
\begin{equation}
	\begin{cases}
		\bm{v}_t = \beta_2 \bm{v}_{t - 1} + (1 - \beta_2) \bm{g}_t^2,
		\\
		\overset{\frown}{\bm{v}}_t = \frac{\bm{v}_t}{1 - \beta_2^t}
	\end{cases}
	\label{eq-som-estimate}
\end{equation}
where $\bm{g}_t$ denotes the gradient of the current iteration, $\bm{v}_{t - 1}$ denotes the second moment of historical iterations. Iterating Equation~\eqref{eq-som-estimate}, we have:
\begin{equation}
	\bm{v}_t = (1 - \beta_2) \sum_{i=1}^{t} \beta_2^{t-i}\bm{g}_i^2
	\label{eq-som-estimate-iter}
\end{equation}
Substituting Equation~\eqref{eq-sgd-gd-noise-S2} into Equation~\eqref{eq-som-estimate-iter} yields:
\begin{equation}
	\begin{aligned}
		\bm{v}_t &= (1 - \beta_2) \sum_{i=1}^{t} \beta_2^{t-i} 
		(\nabla f(\bm{\theta}_{i-1}) + \sigma \bm{\xi})^2
		\\
		&= (1 - \beta_2) \sum_{i=1}^{t} \beta_2^{t-i} 
		(
		(\nabla f(\bm{\theta}_{i-1}))^2 + 
		2\nabla f(\bm{\theta}_{i-1}) \odot \sigma\bm{\xi} +
		\sigma^2 \bm{\xi}^2
		)
	\end{aligned}
	\label{eq-som-estimate-iter-S2}
\end{equation}
Computing the expectation of the second moment estimate:
\begin{equation}
	\mathbb{E}[\bm{v}_t] = (1 - \beta_2^t) \sigma^2 \bm{1} + 
	(1 - \beta_2)\sum_{i=1}^{t} \beta_2^{t-i} \mathbb{E}\left[(\nabla f(\bm{\theta}_{i-1}))^2\right]
	\label{eq-som-estimate-exp}
\end{equation}
Consider the bias term:
\begin{equation}
	\mathbb{E}[\overset{\frown}{\bm{v}}_t] =
	\mathbb{E}\left[\frac{\bm{v}_t}{1 - \beta_2^t}\right] = 
	\sigma^2 \bm{1} + 
	\frac{1 - \beta_2}{1 - \beta_2^t}
	\sum_{i=1}^{t} \beta_2^{t-i} \mathbb{E}\left[(\nabla f(\bm{\theta}_{i-1}))^2\right]
	\label{eq-som-estimate-exp-S2}
\end{equation}
Computing the variance of the second moment estimate:
\begin{equation}
	\begin{aligned}
		\mathrm{Var}[\bm{v}_t] &= \mathbb{E}[\bm{v}_t^2] - \mathbb{E}[\bm{v}_t]^2
		\\
		&= (1 - \beta_2)^2 \sum_{i=1}^{t} \beta_2^{2(t-i)} 
		\left(
		2\sigma^4 \bm{1} +
		4\sigma^2 \mathbb{E}\left[(\nabla f(\bm{\theta}_{i-1}))^2\right] +
		\mathrm{Var}\left[(\nabla f(\bm{\theta}_{i-1}))^2\right]
		\right)
	\end{aligned}
	\label{eq-som-estimate-var}
\end{equation}
Consider the bias term:
\begin{equation}
	\mathrm{Var}[\overset{\frown}{\bm{v}}_t] = \frac{\mathrm{Var}[\bm{v}_t]}{(1 - \beta_2^t)^2}
	\label{eq-som-estimate-var-S2}
\end{equation}
Then, when $t \to \infty$, $\mathrm{Var}[\overset{\frown}{\bm{v}}_t]$ converges to:
\begin{equation}
	\lim\limits_{t \to \infty} \mathrm{Var}[\overset{\frown}{\bm{v}}_t] =
	\frac{1 - \beta_2}{1 + \beta_2} \left(
	2\sigma^4 \bm{1} +
	4\sigma^2 \mathbb{E}\left[(\nabla f(\bm{\theta}_{i-1}))^2\right] + 
	\mathrm{Var}\left[(\nabla f(\bm{\theta}_{i-1}))^2\right]
	\right)
	\label{eq-som-estimate-var-lim}
\end{equation}

Under these assumptions, Equation~\eqref{eq-som-estimate-var-lim} shows that the variance of the second-moment estimate decreases as $\beta_2$ increases during the plateau phase (\emph{i.e.}, $t \to \infty$); as $\beta_2 \to 1$, this variance term tends to zero. This explains why a larger $\beta_2$ can reduce variation in the update scale, but it does not by itself prove that Adam strictly degenerates into momentum SGD. A very large $\beta_2$ also makes the second-moment estimate slower to respond to changes in the gradient distribution, and its bias correction approaches its steady state more slowly. Thus, $\beta_2$ trades smoothing against responsiveness. This is a mechanism explanation under the stated assumptions, not a nonconvex convergence-rate, regret, or general-stability result.

\subsection{Novel exponential decay rate for the second-order moment estimation and AdamNX}\label{sec3.2}

\begin{figure}[H]
	\centering
	\includegraphics[width=0.8\linewidth]{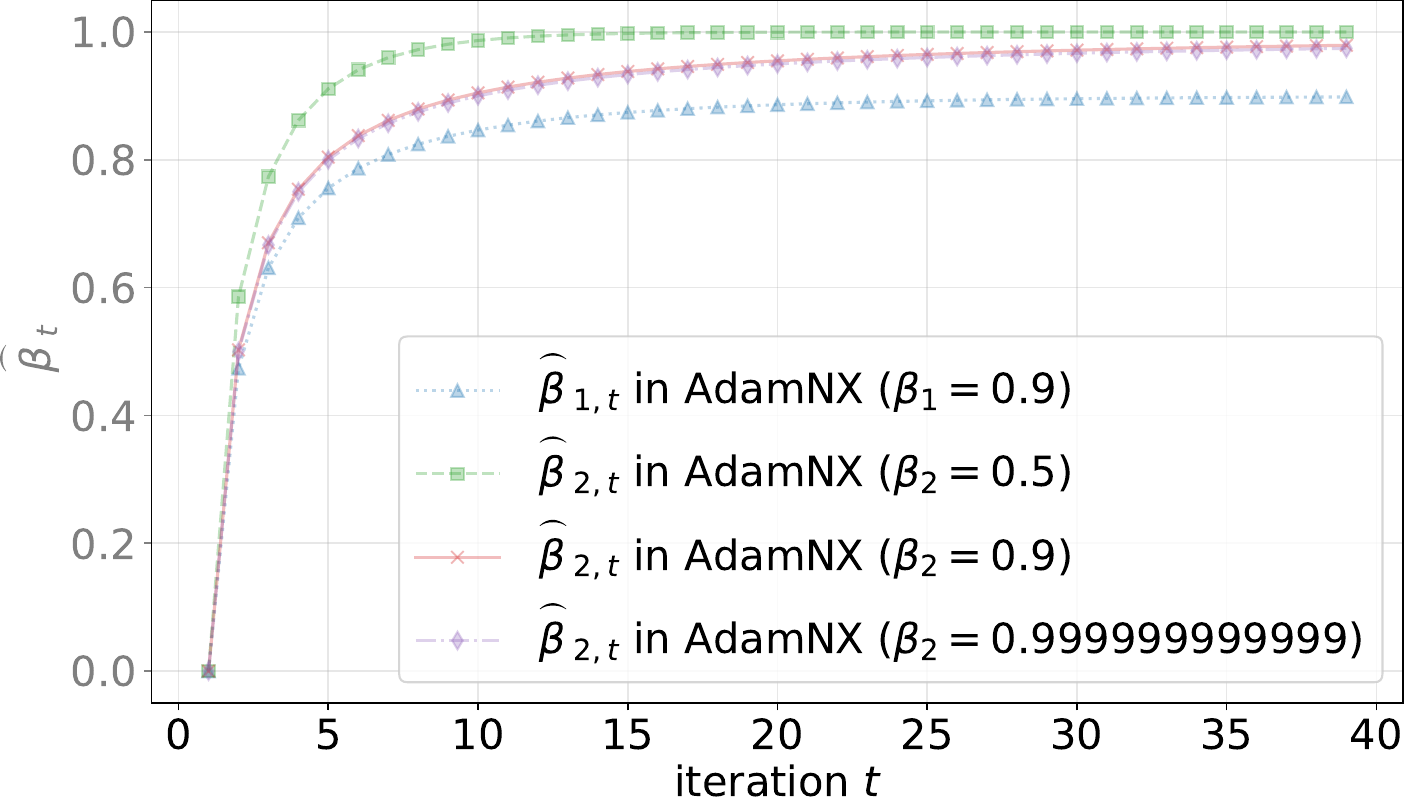}%
	\caption{Functional curves of $\overset{\frown}{\beta}_{1,t}$ and $\overset{\frown}{\beta}_{2,t}$ versus iteration number $t$ in AdamNX}
	\label{fig-adamnx-beta-t}
\end{figure}

For clarity, $\beta_1$ and $\beta_2$ are constant base decay rates, while $\overset{\frown}{\beta}_{1,t}$ and $\overset{\frown}{\beta}_{2,t}$ are time-varying coefficients. The symbol $\bm{v}_t$ denotes the raw second-moment estimate and $\overset{\frown}{\bm{v}}_t$ its bias-corrected form. We consider a $\beta_2$ that varies with iteration $t$, with a corresponding adjustment of $\beta_1$, and has the following properties:
\begin{enumerate}
	\item[(1)] $\overset{\frown}{\beta}_{2,t=1} = 0$, $\overset{\frown}{\beta}_{2,t \to \infty} = 1$.
	\item[(2)] For all $t \ge 1$, having: $\overset{\frown}{\beta}_{2,t} \ge \overset{\frown}{\beta}_{1,t}$.
	\item[(3)] During the early and middle stages of training, $\overset{\frown}{\beta}_{2,t}$ is as close as possible to $\overset{\frown}{\beta}_{1,t}$.
\end{enumerate}
Therefore, this paper proposes a novel decay rate for the second moment estimate:
\begin{equation}
	\overset{\frown}{\beta}_{2,t} = 
	\frac{1 - \beta_2^{(1-\beta_2)(t-1)}}{1 - \beta_2^{(1-\beta_2)t}}
	\label{eq-adamx-beta2t}
\end{equation}
It is easy to prove that:
\begin{equation}
	\lim\limits_{t \to \infty} \overset{\frown}{\beta}_{2,t} = 
	\lim\limits_{t \to \infty} 
	\frac{1 - \beta_2^{(1-\beta_2)(t-1)}}{1 - \beta_2^{(1-\beta_2)t}}
	= 1
	\label{eq-adamx-beta2t-lim}
\end{equation}
and correspondingly proposes the AdamNX algorithm, as shown in Algorithm~\ref{algo-adamnx}. To align with Adam's default settings: $\beta_1 = 0.9$, $\beta_2 = 0.999$, AdamNX is set to: $\beta_1 = 0.9$, $\beta_2 = 0.99$. This allows hyperparameters tuned for Adam to be better transferred to AdamNX. Figure~\ref{fig-adamnx-beta-t} visualizes $\overset{\frown}{\beta}_{1,t}$ and $\overset{\frown}{\beta}_{2,t}$ in AdamNX.

\begin{algorithm}[H]
	\caption{AdamNX algorithm}
	\label{algo-adamnx}
	\MyKwIn[Input$_1$]{training dataset $\bm{\mathcal{D}}$.}
	\MyKwIn[Input$_2$]{parameters to be optimized $(\bm{\theta}_1, \bm{\theta}_2, \dots, \bm{\theta}_L)$, first-order moments $(\bm{m}_1, \bm{m}_2, \dots, \bm{m}_L)$, second-order moments $(\bm{v}_1, \bm{v}_2, \dots, \bm{v}_L)$.}
	\MyKwIn[Input$_3$]{learning rate $\eta_t$, weight decay rate $\lambda$, exponential decay rate $(\beta_1, \beta_2) = (0.9, 0.99)$, $\epsilon = 1 \times 10^{-8}$, total number of iterations $T$.}
	\MyKwOut{extremum $(\bm{\theta}_1^*, \bm{\theta}_2^*, \dots, \bm{\theta}_L^*)$.}
	\BlankLine
	\For{$t = 1$ \KwTo $T$}{
		randomly drawing a subset of training samples: $\bm{\mathcal{R}}_t \subseteq \bm{\mathcal{D}}$\;
		\tcc{forward propagation}
		loss value calculation: $\mathcal{L}_t = \frac{1}{\lvert\bm{\mathcal{R}}_t\rvert}\sum_{(\bm{x}_i, \bm{y}_i) \in \bm{\mathcal{R}}_t}
		E(\overset{\frown}{\bm{y}}(\bm{x}_i, \bm{\theta}_{1,t - 1}, \dots, \bm{\theta}_{L,t-1}), \bm{y}_i)$\;
		\tcc{backpropagation}
		\For{$l = 1$ \KwTo $L$}{
			gradient calculation: $\bm{g}_{l,t} = \nabla_{\bm{\theta}_{l,t-1}}\mathcal{L}$\;
			first-order moment estimate: $\bm{m}_{l,t} = \frac{\beta_1 - \beta_1^t}{1 - \beta_1^t} \bm{m}_{l,t-1} + 
			\left(1 - \frac{\beta_1 - \beta_1^t}{1 - \beta_1^t}\right) \bm{g}_{l,t}$\;
			second-order moment estimate: $\bm{v}_{l,t} = \frac{1 - \beta_2^{(1-\beta_2)(t-1)}}{1 - \beta_2^{(1-\beta_2)t}} \bm{v}_{l,t-1} + 
			\left(1 - \frac{1 - \beta_2^{(1-\beta_2)(t-1)}}{1 - \beta_2^{(1-\beta_2)t}}\right) \bm{g}_{l,t}^2$\;
			gradient element normalization: $\bm{u}_{l,t} = \frac{\bm{m}_{l,t}}{\sqrt{\bm{v}_{l,t}} + \epsilon}$\;
			parameter updating: $\bm{\theta}_{l,t} = \bm{\theta}_{l,t-1} - 
			\eta_t \left(\bm{u}_{l,t} + \lambda_l \bm{\theta}_{l,t-1}\right)$, \newline 
			$\lambda_l = \begin{cases}
				\lambda, &\text{if }\bm{\theta}_l\text{ is a matrix parameter} \\
				0, &\text{if }\bm{\theta}_l\text{ is not a matrix parameter} 
			\end{cases}$\;
		}
		\tcc{saving the optimal loss value and extremum point}
		\If{$\mathcal{L}_t < \mathcal{L}_0$}{
			$\mathcal{L}_0 \gets \mathcal{L}_t$\;
			\For{$l = 1$ \KwTo $L$}{
				$\bm{\theta}_l^* \gets \bm{\theta}_{l,t}$\tcp*{storing the extremum point}
			}
		}
	}
\end{algorithm}

\subsection{AdamNX \emph{v.s.} Adam}\label{sec3.3}

\begin{figure}[H]
	\centering
	\includegraphics[width=0.8\linewidth]{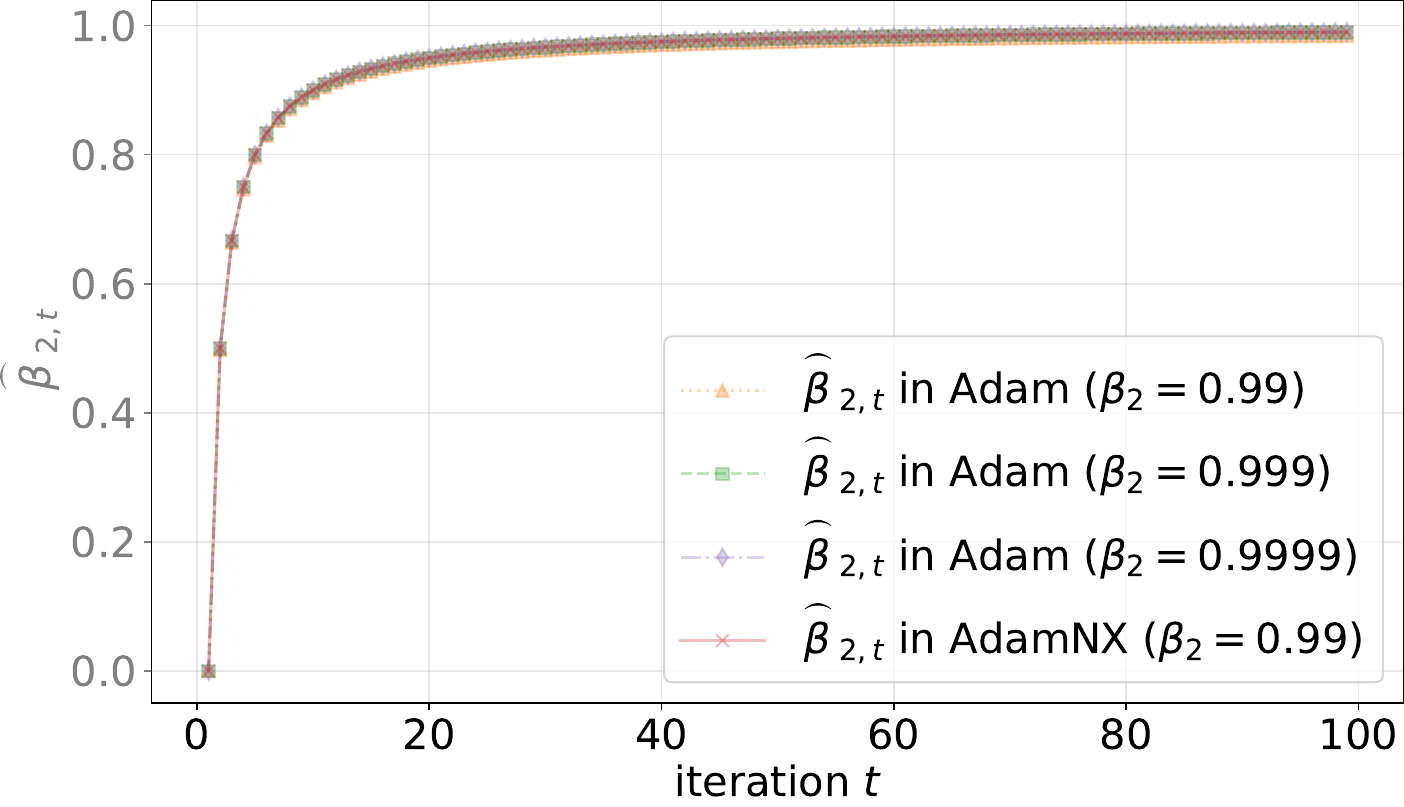}%
	\caption{Comparison between AdamNX's $\overset{\frown}{\beta}_{2,t}$ and Adam's $\overset{\frown}{\beta}_{2,t}$}
	\label{fig-adam-adamnx-beta-t}
\end{figure}

In fact, the recurrence of Adam's bias-corrected $\overset{\frown}{\bm{v}}_t$ also implicitly contains $\overset{\frown}{\beta}_{2,t}$:
\begin{equation}
	\begin{aligned}
		\overset{\frown}{\bm{v}}_t 
		&= \frac{\bm{v}_t}{1 - \beta_2^t}
		= \frac{\beta_2 \bm{v}_{t - 1} + (1 - \beta_2) \bm{g}_t^2}{1 - \beta_2^t}
		=  \frac{(\beta_2 - \beta_2^t) \overset{\frown}{\bm{v}}_{t - 1} + (1 - \beta_2) \bm{g}_t^2}{1 - \beta_2^t}
		\\
		&= \frac{\beta_2 - \beta_2^t}{1 - \beta_2^t} \overset{\frown}{\bm{v}}_{t - 1} + 
		\left(1 - \frac{\beta_2 - \beta_2^t}{1 - \beta_2^t} \right) \bm{g}_t^2
		\triangleq \overset{\frown}{\beta}_{2,t} \bm{v}_{t - 1} + (1 - \overset{\frown}{\beta}_{2,t}) \bm{g}_t^2
	\end{aligned}
	\label{eq-som-estimate-bias-correction}
\end{equation}
And it is easy to prove:
\begin{equation}
	\lim\limits_{t \to \infty} \frac{\beta_2 - \beta_2^t}{1 - \beta_2^t} = \beta_2
	\label{eq-adam-beta2t-lim}
\end{equation}
Comparing Equations~\eqref{eq-adamx-beta2t-lim} and \eqref{eq-adam-beta2t-lim}, Adam's corresponding coefficient tends to $\beta_2$ as $t \to \infty$, whereas the AdamNX coefficient tends to $1$. Within the recurrence interpretation and assumptions used here, this makes AdamNX updates more momentum-SGD-like late in training. This describes a mechanism trend rather than strict degeneration or general stability. Figure~\ref{fig-adam-adamnx-beta-t} further visualizes the comparison between AdamNX's $\overset{\frown}{\beta}_{2,t}$ and Adam's $\overset{\frown}{\beta}_{2,t}$ for different $\beta_2$ values.

\section{Experimental reproduction details}\label{sec4}

This section introduces the common experimental details for Sections~\ref{sec5} and \ref{sec6}. We report single results for the visual tasks, configurations, and comparators available in this paper. We do not report multi-seed means, standard deviations, significance tests, Hessian or sharpness measurements, timing or memory measurements, a unified tuning budget, or non-visual tasks. The results below therefore describe comparisons within these reported configurations and do not support claims of statistical optimality, engineering efficiency, or cross-task generalization.

Table~\ref{tbl-software-hardware-configuration} shows the main software and hardware configurations required for the experiments.

\begin{table}[H]
	\scriptsize
	\caption{Main software and hardware configurations}
	\label{tbl-software-hardware-configuration}
	\begin{tblr}{
			width=\linewidth, 
			colspec={X[l] X[l]}, 			 
			hline{1,Z}={1pt},                    
			columns={valign=m},                  
			rowsep=0.5pt                         
		}
		Torch~\cite{2019-Paszke-PyTorch} & 2.8.0 \\
		Torchvision~\cite{2019-Paszke-PyTorch} & 0.23.0 \\
		Python & 3.13.7 \\
		NVIDIA CUDA Toolkit & 12.9 \\
		NVIDIA cuDNN & 10.2 \\
		Memory Capacity & 192 GB \\
		CPU & Intel(R) Core i9-14900KF \\
		GPU & NVIDIA RTX A5000 \\
	\end{tblr}
\end{table}

\begin{figure}[H]
	{\centering
		\includegraphics[width=0.8\linewidth]{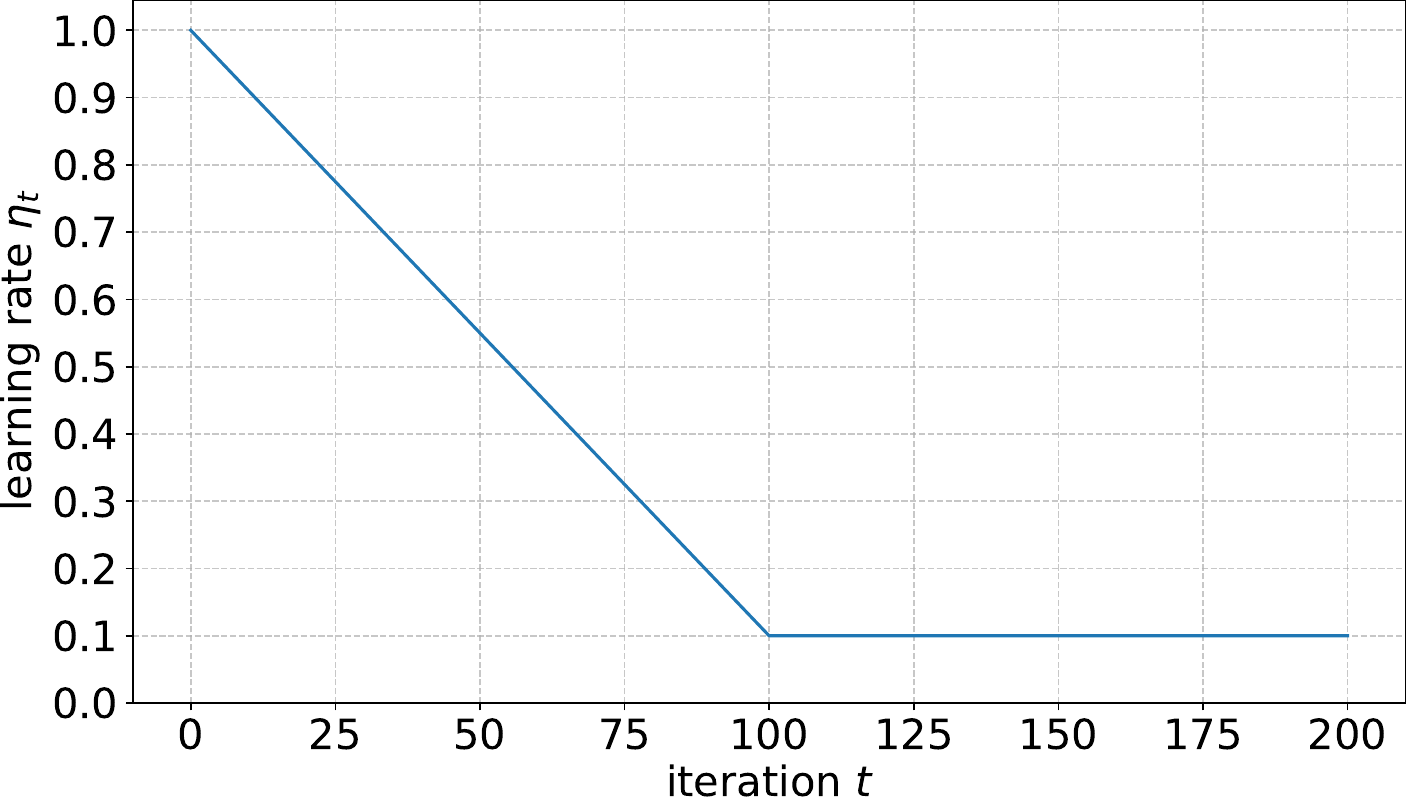}%
		\caption{Functional curve between the learning rate and the number of iterations}
		\label{fig-lr-strategy}}
	\footnotesize{Note: The settings of the horizontal and vertical coordinates in the figure are only for intuitively presenting the trend of the learning rate changing with the iteration process, and the values do not reflect the actual data in the experiments.}  
\end{figure}

The learning rate strategy configuration. The experiments in the following sections followed two different learning rate strategies depending on the model. One was a fixed learning rate, and the other followed the learning rate strategy shown in Equation~\eqref{eq-lr-strategy}:
\begin{equation}
	\eta_t = \begin{cases}
		\eta_{\mathrm{peak}} + \frac{\eta_{\min} - \eta_{\mathrm{peak}}}{t_1} t, &\text{if } 0 \le t \le t_1, \\
		\eta_{\min}, &\text{if } t > t_1
	\end{cases}
	\label{eq-lr-strategy}
\end{equation}
where $\eta_{\min}$ denotes the minimum learning rate, and $\eta_{\mathrm{peak}}$ denotes the peak learning rate. Figure~\ref{fig-lr-strategy} visualizes the functional relationship between the learning rate and the number of iterations.

\section{Comparison results and analysis of optimization algorithms of the same type}\label{sec5}

This section compares AdamNX with Adam~\cite{2015-Kingma-Adam} and AdaX~\cite{2020-Li-Adax}. The experiments cover CIFAR-100~\cite{2009-Krizhevsky-CIFAR-100} for image classification, PASCAL VOC~\cite{2011-Everingham-VOC2012} for object detection, and Semantic Boundaries~\cite{2011-Hariharan-SBD} for semantic segmentation, using EfficientNetV2~\cite{2021-Tan-EfficientNetV2}, SwinV2~\cite{2022-Liu-SwinV2}, YOLOv7-tiny~\cite{2023-Wang-YOLOv7}, and U-Net~\cite{2015-Ronneberger-U-Net}. These comparisons do not include AdamW, SWATS, or AdaI.

\subsection{Image classification experimental results and analysis}\label{sec5.1}

Benchmark dataset for image classification. CIFAR-100~\cite{2009-Krizhevsky-CIFAR-100} is a benchmark image classification dataset constructed by the Department of Computer Science at the University of Toronto in 2009. As an extended version of the CIFAR series, it contains $60,000$ RGB color images with $32 \times 32$ pixel, forming a $100$-class fine-grained classification system with $20$ superclasses (each containing $5$ subclasses). This dataset is strictly divided into $50,000$ training images and $10,000$ test images. During training, data augmentation strategies including random color space transformation and random horizontal flipping were employed.

Benchmark models for image classification. EfficientNetV2~\cite{2021-Tan-EfficientNetV2} is a convolutional neural network (CNN) series proposed by Google. Its core innovation lies in the compound scaling method, which uniformly adjusts the network depth, width, and input resolution to achieve the optimal balance between model efficiency and performance. SwinV2\cite{2022-Liu-SwinV2} is an improved version of the visual Transformer proposed by Microsoft, with several innovations aimed at large-scale model training. To match the image resolution in CIFAR, EfficientNetV2 retained only the last three downsamplings; in SwinV2, the patch size was set to $1$, and the window size was set to $4$. The reproduction of all models directly cloned the official implementation of torchvision, and model weight initialization followed the official default without loading pre-trained weights.

Evaluation metrics for image classification. Top-1 error rate and Top-5 error rate are mainstream metrics for measuring the performance of image classification models. Suppose the test set contains $N$ images, where the true label of the $i$-th image is $y_i$, and the model outputs a class probability vector for this image denoted as $p_i = (p_{i,1}, p_{i,2}, \dots, p_{i,C})$ ($C$ is the total number of classes), then the Top-1 error rate and Top-5 error rate are defined as:
\begin{align}
	&E_{\mathrm{Top}\mbox{-}1} = \frac{1}{N}\sum_{i=1}^{N}
	\chi_A\left(y_i \neq \arg\max_{j} p_{i,j}\right)
	\\
	&E_{\mathrm{Top}\mbox{-}5} = \frac{1}{N}\sum_{i=1}^{N}\chi_A
	\left(y_i \notin \mathrm{top}\mbox{-}5\left({p_{i,j}}_{j=1}^{C}\right)\right)
\end{align}
where $\chi_A(.)$ denotes the indicator function. The above definitions follow the convention of reference~\cite{2012-Krizhevsky-ImageNetClassification} and are widely used in subsequent studies to unify the evaluation criteria.

Lowest error rate comparison. Table~\ref{tbl-cifar100-EfficientNetV2-M} shows the comparative experimental results on the benchmark model EfficientNetV2-M. The total number of training epochs was set to $80$, the batch size was set to $64$, the peak learning rate was set to $3 \times 10^{-3}$, the minimum learning rate was set to $3 \times 10^{-5}$, and the training $t_1$ was set to $37,488$. As seen in Table~\ref{tbl-cifar100-EfficientNetV2-M}, our AdamNX achieved the lowest top-1 error rate of $34.27\%$ and the lowest top-5 error rate of $11.40\%$ on EfficientNetV2-M. Table~\ref{tbl-cifar100-SwinV2-S} shows the comparative experimental results on the benchmark model SwinV2-S. The total number of training epochs was set to $80$, the batch size was set to $64$, and the fixed learning rate was set to $2.2 \times 10^{-5}$. As seen in Table~\ref{tbl-cifar100-SwinV2-S}, our AdamNX also achieved the lowest top-1 error rate of $39.42\%$ and the lowest top-5 error rate of $14.68\%$ on SwinV2-S.

\begin{table}[H]
	\scriptsize
	\caption{Image classification results on CIFAR-100 for optimization algorithms of the same type (benchmark model is EfficientNetV2-M)}
	\label{tbl-cifar100-EfficientNetV2-M}
	\begin{threeparttable}
		\begin{tblr}{
				width=\linewidth, 
				colspec={X[l] X[c] X[c]}, 	
				hline{1,Z}={1pt},                    
				hline{2}={0.5pt},                 	 
				columns={valign=m},                  
				rowsep=0.5pt                         
			}
			& Top-1 err. ($\%$) $\downarrow$ 
			& Top-5 err. ($\%$) $\downarrow$ \\
			Adam~\cite{2015-Kingma-Adam}  & 39.44 & 14.76 \\
			AdaX~\cite{2020-Li-Adax}  & 36.35 & 12.99 \\
			AdamNX (Ours) & \textbf{34.27} & \textbf{11.40} \\
		\end{tblr}
		\begin{tablenotes}  
			\item Note: the \textbf{bold} denotes the best results in each column.
		\end{tablenotes}
	\end{threeparttable}
\end{table}

\begin{table}[H]
	\scriptsize
	\caption{Image classification results on CIFAR-100 for optimization algorithms of the same type (benchmark model is SwinV2-S)}
	\label{tbl-cifar100-SwinV2-S}
	\begin{tblr}{
			width=\linewidth, 
			colspec={X[l] X[c] X[c]}, 	
			hline{1,Z}={1pt},                    
			hline{2}={0.5pt},                 	 
			columns={valign=m},                  
			rowsep=0.5pt                         
		}
		& Top-1 err. ($\%$) $\downarrow$ 
		& Top-5 err. ($\%$) $\downarrow$ \\
		Adam~\cite{2015-Kingma-Adam}  & 40.05 & 14.80 \\
		AdaX~\cite{2020-Li-Adax}  & 39.97 & 15.07 \\
		AdamNX (Ours) & \textbf{39.42} & \textbf{14.68} \\
	\end{tblr}
\end{table}

\subsection{Object detection experimental results and analysis}\label{sec5.2}

Object detection benchmark datasets. Following the standard protocol in the object detection field of PASCAL VOC~\cite{2011-Everingham-VOC2012}: the trainval subsets of VOC2007 and VOC2012 were merged (a total of $16,551$ images containing $20$ classes of objects) as the training data, and the publicly annotated VOC2007 test ($4,952$ images) was used for model evaluation. During training, data augmentation strategies including random padding and scaling, random color space transformation, and random horizontal flipping were employed.

Benchmark model for object detection. YOLOv7-tiny was the high-speed lightweight version of the YOLOv7~\cite{2023-Wang-YOLOv7} series. It achieved millisecond-level inference while maintaining high detection accuracy through a streamlined ELAN-Tiny backbone network, balancing detection accuracy and efficiency. The image resolution fed into the model was set to $320 \times 320$. Model weight initialization followed the initialization strategy of RegNet~\cite{2020-Radosavovic-RegNet}.

Evaluation metrics for object detection. Mean Average Precision (mAP) is the most commonly used evaluation metric in the field of object detection, and its calculation process is as follows. For each class $j$, all prediction boxes are sorted in descending order of confidence and matched with the ground-truth boxes one by one. If $\mathrm{IoU} \ge T$ (\emph{e.g.}, $T=0.5$ or $0.75$), and the classes are consistent, it is recorded as a True Positive (TP); otherwise, it is a False Positive (FP). Precision and recall are cumulatively calculated as follows:
\begin{align}
	&P_j(k) = \frac{\sum_{i=1}^{k} \mathrm{TP}_i}{k}
	\label{eq-precision}
	\\
	&R_j(k) = \frac{\sum_{i=1}^{k} \mathrm{TP}_i}{N_j}
	\label{eq-recall}
\end{align}
where $N_j$ denotes the total number of ground-truth boxes for class $j$, and $k$ denotes the top $k$ prediction boxes. The $P_j(R_j)$ curve is plotted, and the area under the curve is taken to obtain the average precision (AP) for this class:
\begin{equation}
	\mathrm{AP}_j = \int_{0}^{1} P_j(R_j) \mathrm{d} R_j
	\label{eq-ap}
\end{equation}
The mean value over all classes is then calculated as:
$
\mathrm{mAP} = \frac{1}{C} \sum_{j=1}^{C} \mathrm{AP}_j
$.

Highest mAP comparison. Table~\ref{tbl-VOC2007} shows the comparative experimental results on the VOC2007 test benchmark dataset. The total number of training epochs was set to $80$, the batch size was set to $32$, the peak learning rate was set to $3 \times 10^{-3}$, the minimum learning rate was set to $3 \times 10^{-5}$, and the training $t_1$ was set to $25,680$. As seen in Table~\ref{tbl-VOC2007}, our AdamNX achieved the highest mAP@0.5 of $52.18\%$ and the highest mAP@0.75 of $28.21\%$.

\begin{table}[H]
	\scriptsize
	\caption{Object detection results on VOC2007 test for optimization algorithms of the same type}
	\label{tbl-VOC2007}
	\begin{threeparttable}
		\begin{tblr}{
				width=\linewidth, 
				colspec={X[l] X[c] X[c]}, 	
				hline{1,Z}={1pt},                    
				hline{2}={0.5pt},                 	 
				columns={valign=m},                  
				rowsep=0.5pt                         
			}
			& mAP@0.5 ($\%$) $\uparrow$ 
			& mAP@0.75 ($\%$) $\uparrow$ \\
			Adam~\cite{2015-Kingma-Adam}  & 52.07 & 27.45 \\		
			AdaX~\cite{2020-Li-Adax}  & 51.99 & 27.19 \\
			AdamNX (Ours) & \textbf{52.18} & \textbf{28.21} \\
		\end{tblr}
		\begin{tablenotes}  
			\item Note: mAP@0.5 denotes $\mathrm{IoU} \ge 0.5$, and mAP@0.75 denotes $\mathrm{IoU} \ge 0.75$.
		\end{tablenotes}
	\end{threeparttable}
\end{table}

\subsection{Semantic segmentation experimental results and analysis}\label{sec5.3}

Benchmark dataset for semantic segmentation. The Semantic Boundaries~\cite{2011-Hariharan-SBD} dataset contains $11,355$ images from the PASCAL VOC 2011 dataset, divided into $8,498$ training images and $2,857$ validation images, providing pixel-level semantic segmentation annotations and object boundary information for $21$ classes of objects, supporting semantic segmentation and boundary detection tasks. During training, data augmentation strategies including random scaling, random horizontal flipping, and random cropping were employed.

Benchmark model for semantic segmentation. U-Net~\cite{2015-Ronneberger-U-Net} was initially proposed for biomedical image segmentation. Its symmetric encoder-decoder structure integrates high-resolution details and low-resolution semantic information through skip connections, enabling pixel-level accurate prediction with limited annotated data. Given its design that balances localization accuracy and context modeling capabilities, U-Net has become a widely adopted benchmark model in the field of semantic segmentation. The parameter basic units in U-Net were set to $32$, and the image resolution fed into the model is set to $320 \times 320$. Model weight initialization followed the initialization strategy of RegNet~\cite{2020-Radosavovic-RegNet}. The loss function employed a weighted average combination of cross-entropy loss~\cite{1986-Rumelhart-GD-SGD} and Dice loss~\cite{2016-Milletari-DiceLoss}.

Evaluation metric for semantic segmentation. Mean Intersection over Union (mIoU) is the most commonly used evaluation metric in the field of semantic segmentation, and its calculation method is as follows: for each class, the intersection over union between the prediction results and the ground truth is calculated and then averaged:
\begin{equation}
	\mathrm{mIoU} = \frac{1}{C} \sum_{i=1}^{C} \frac{\mathrm{TP}_i}{\mathrm{TP}_i + \mathrm{FP}_i + \mathrm{FN}_i}
	\label{eq-mIoU}
\end{equation}
where $C$ denotes the total number of classes, and $\mathrm{TP}_i$, $\mathrm{FP}_i$, $\mathrm{FN}_i$ denote the number of TP, FP, and false negative pixels for class $i$, respectively. This metric measures both localization accuracy and classification accuracy simultaneously, and a higher value indicates better segmentation performance.

Highest mIoU comparison. Table~\ref{tbl-SBD} shows the comparative experimental results on the Semantic Boundaries benchmark dataset. The total number of training epochs was set to $80$, the batch size was set to $32$, the peak learning rate was set to $1 \times 10^{-3}$, the minimum learning rate was set to $1 \times 10^{-5}$, and the training $t_1$ was set to $12,720$. As can be seen from Table~\ref{tbl-SBD}, our AdamNX achieved the highest mIoU of $37.81\%$.

\begin{table}[H]
	\scriptsize
	\caption{Semantic segmentation results on Semantic Boundaries for optimization algorithms of the same type}
	\label{tbl-SBD}
	\begin{tblr}{
			width=\linewidth, 
			colspec={X[l] X[c]}, 	
			hline{1,Z}={1pt},                    
			hline{2}={0.5pt},                 	 
			columns={valign=m},                  
			rowsep=0.5pt                         
		}
		& mIoU ($\%$) $\uparrow$ \\
		Adam~\cite{2015-Kingma-Adam}  & 34.43 \\
		AdaX~\cite{2020-Li-Adax}  & 37.23 \\
		AdamNX (Ours) & \textbf{37.81} \\
	\end{tblr}
\end{table}

\subsection{Ablation study results and analysis}\label{sec5.4}

\begin{figure}[H]
	\centering
	\includegraphics[width=0.8\linewidth]{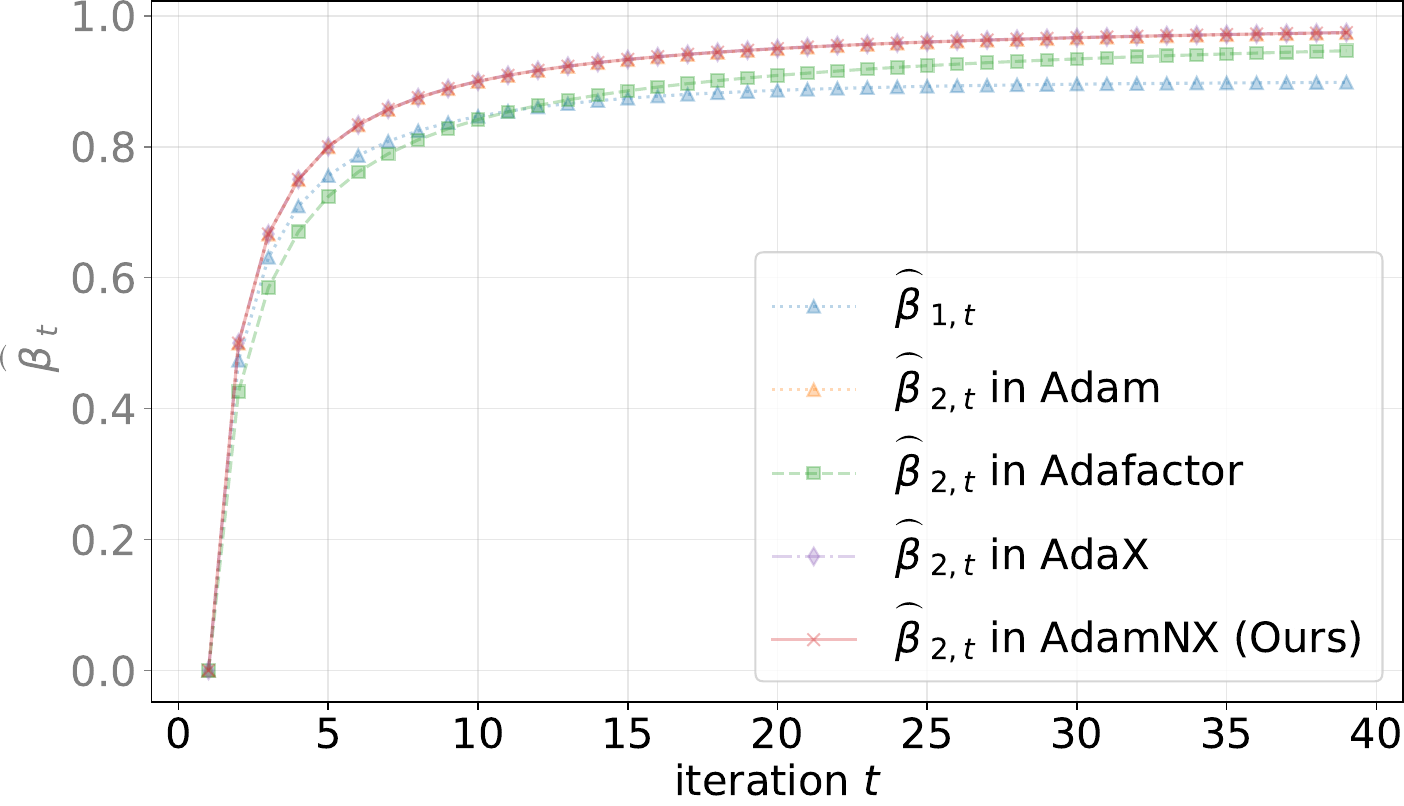}%
	\caption{Different second-order moment estimate exponential decay rates}
	\label{fig-beta-t}
\end{figure}

Figure~\ref{fig-beta-t} visualizes the second-order moment estimate exponential decay rates in different optimization algorithms. The ablation experiments in this section focused on comparing the different second-order moment estimate exponential decay rates shown in this figure. Therefore, the $\overset{\frown}{\beta}_{2,t}$ in Adam was replaced with the $\overset{\frown}{\beta}_{2,t}$ from Adafactor~\cite{2018-Shazeer-Adafactor}, AdaX, and AdamNX, respectively, to verify which second-order moment estimate exponential decay rate is superior.

Reported-result comparison. Table~\ref{tbl-som-decay-rate-abalation-study-cifar100-EfficientNetV2-M} shows the ablation results for EfficientNetV2-M on CIFAR-100. Table~\ref{tbl-som-decay-rate-abalation-study-cifar100-SwinV2-S} reports the corresponding results for SwinV2-S, Table~\ref{tbl-som-decay-rate-abalation-study-VOC2007} for YOLOv7-tiny on VOC2007 test, and Table~\ref{tbl-som-decay-rate-abalation-study-SBD} for U-Net on Semantic Boundaries. The AdamNX coefficient obtains the tabulated values in these configurations. Single reported results do not establish statistical optimality; the following are analytic properties of the coefficient, not properties demonstrated by these experiments:
\begin{itemize}
	\item[(1)] $\overset{\frown}{\beta}_{2,t=1} = 0$, $\overset{\frown}{\beta}_{2,t \to \infty} = 1$.
	\item[(2)] For all $t \ge 1$, it holds that: $\overset{\frown}{\beta}_{2,t} \ge \overset{\frown}{\beta}_{1,t}$.
\end{itemize}

\begin{table}[H]
	\scriptsize
	\caption{Image classification ablation experimental results for second-order moment estimate exponential decay rates (benchmark model is EfficientNetV2-M)}
	\label{tbl-som-decay-rate-abalation-study-cifar100-EfficientNetV2-M}
	\begin{tblr}{
			width=\linewidth, 
			colspec={X[2,l] X[c] X[c]}, 	
			hline{1,Z}={1pt},                    
			hline{2}={0.5pt},                 	 
			columns={valign=m},                  
			rowsep=0.5pt                         
		}
		Adam\cite{2015-Kingma-Adam} + $\overset{\frown}{\beta}_{2,t}$ 
		& Top-1 err. ($\%$) $\downarrow$ 
		& Top-5 err. ($\%$) $\downarrow$ \\
		$(\beta_2 - \beta_2^t) / (1 - \beta_2^t)$, $\beta_2 = 0.999$ (Adam~\cite{2015-Kingma-Adam})
		& 39.44
		& 14.76 \\
		$1 - 1 / t^c$, $c = 0.8$ (Adafactor~\cite{2018-Shazeer-Adafactor})  
		& 41.22
		& 15.23 \\
		$1 - \beta_2 / ((1 + \beta_2)^t - 1)$, $\beta_2 = 0.0001$ (AdaX~\cite{2020-Li-Adax})
		& 35.22
		& 11.92 \\
		$\left(1 - \beta_2^{(1-\beta_2)(t-1)}\right) / \left(1 - \beta_2^{(1-\beta_2)t}\right)$, $\beta_2 = 0.99$ (Ours) 
		& \textbf{34.27}
		& \textbf{11.40} \\
	\end{tblr}
\end{table}

\begin{table}[H]
	\scriptsize
	\caption{Image classification ablation experimental results for second-order moment estimate exponential decay rates (benchmark model is SwinV2-S)}
	\label{tbl-som-decay-rate-abalation-study-cifar100-SwinV2-S}
	\begin{threeparttable}
		\begin{tblr}{
				width=\linewidth, 
				colspec={X[2,l] X[c] X[c]}, 	
				hline{1,Z}={1pt},                    
				hline{2}={0.5pt},                 	 
				columns={valign=m},                  
				rowsep=0.5pt                         
			}
			Adam\cite{2015-Kingma-Adam} + $\overset{\frown}{\beta}_{2,t}$ 
			& Top-1 err. ($\%$) $\downarrow$ 
			& Top-5 err. ($\%$) $\downarrow$ \\
			$(\beta_2 - \beta_2^t) / (1 - \beta_2^t)$, $\beta_2 = 0.999$ (Adam~\cite{2015-Kingma-Adam})
			& 40.05
			& 14.80 \\
			$1 - 1 / t^c$, $c = 0.8$ (Adafactor~\cite{2018-Shazeer-Adafactor})
			& 40.64
			&  15.05 \\
			$1 - \beta_2 / ((1 + \beta_2)^t - 1)$, $\beta_2 = 0.0001$ (AdaX~\cite{2020-Li-Adax})
			& 39.64
			& \textbf{14.50} \\
			$\left(1 - \beta_2^{(1-\beta_2)(t-1)}\right) / \left(1 - \beta_2^{(1-\beta_2)t}\right)$, $\beta_2 = 0.99$ (Ours) 
			& \textbf{39.42}
			& \underline{14.68} \\
		\end{tblr}
		\begin{tablenotes}  
			\item Note: the \underline{underline} indicates the second-best results in each column.
		\end{tablenotes}
	\end{threeparttable}
\end{table}

\begin{table}[H]
	\scriptsize
	\caption{Object detection ablation experimental results for second-order moment estimate exponential decay rates}
	\label{tbl-som-decay-rate-abalation-study-VOC2007}
	\begin{tblr}{
			width=\linewidth, 
			colspec={X[2,l] X[c] X[c]}, 	
			hline{1,Z}={1pt},                    
			hline{2}={0.5pt},                 	 
			columns={valign=m},                  
			rowsep=0.5pt                         
		}
		Adam\cite{2015-Kingma-Adam} + $\overset{\frown}{\beta}_{2,t}$ 
		& mAP@0.5 ($\%$) $\uparrow$ 
		& mAP@0.75 ($\%$) $\uparrow$ \\
		$(\beta_2 - \beta_2^t) / (1 - \beta_2^t)$, $\beta_2 = 0.999$ (Adam~\cite{2015-Kingma-Adam})  & 52.07 & 27.45 \\
		$1 - 1 / t^c$, $c = 0.8$ (Adafactor~\cite{2018-Shazeer-Adafactor})   & 45.50 & 21.63 \\
		$1 - \beta_2 / ((1 + \beta_2)^t - 1)$, $\beta_2 = 0.0001$ (AdaX~\cite{2020-Li-Adax})  & 51.74 & 27.24 \\
		$\left(1 - \beta_2^{(1-\beta_2)(t-1)}\right) / \left(1 - \beta_2^{(1-\beta_2)t}\right)$, $\beta_2 = 0.99$ (Ours) & \textbf{52.18} & \textbf{28.21} \\
	\end{tblr}
\end{table}

\begin{table}[H]
	\scriptsize
	\caption{Semantic segmentation ablation experimental results for second-order moment estimate exponential decay rates}
	\label{tbl-som-decay-rate-abalation-study-SBD}
	\begin{tblr}{
			width=\linewidth, 
			colspec={X[l] X[c]}, 	
			hline{1,Z}={1pt},                    
			hline{2}={0.5pt},                 	 
			columns={valign=m},                  
			rowsep=0.5pt                         
		}
		Adam\cite{2015-Kingma-Adam} + $\overset{\frown}{\beta}_{2,t}$ & mIoU ($\%$) $\uparrow$ \\
		$(\beta_2 - \beta_2^t) / (1 - \beta_2^t)$, $\beta_2 = 0.999$ (Adam~\cite{2015-Kingma-Adam})  & 34.43 \\
		$1 - 1 / t^c$, $c = 0.8$ (Adafactor~\cite{2018-Shazeer-Adafactor})  & 31.76 \\
		$1 - \beta_2 / ((1 + \beta_2)^t - 1)$, $\beta_2 = 0.0001$ (AdaX~\cite{2020-Li-Adax}) & 36.37 \\
		$\left(1 - \beta_2^{(1-\beta_2)(t-1)}\right) / \left(1 - \beta_2^{(1-\beta_2)t}\right)$, $\beta_2 = 0.99$ (Ours) & \textbf{37.81} \\
	\end{tblr}
\end{table}

\begin{figure}[H]
	\centering
	\includegraphics[width=0.8\linewidth]{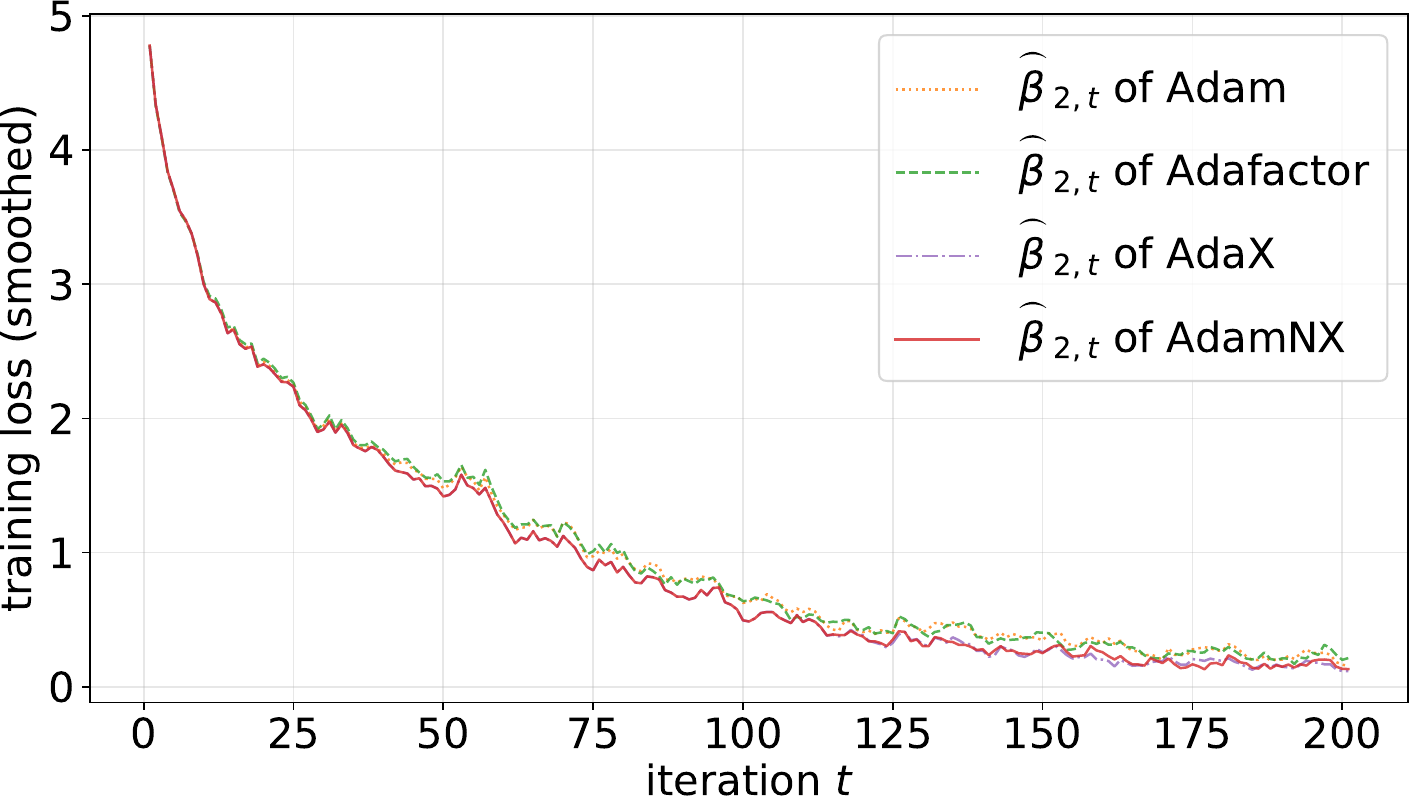}%
	\caption{Training loss \emph{vs.} iteration (experimental logs from Table~\ref{tbl-som-decay-rate-abalation-study-cifar100-SwinV2-S})}
	\label{fig-SOM-cifar100-swinv2-train-loss-vs-iter-step312}
\end{figure}

Comparison of image classification training curves. The experimental logs from Table~\ref{tbl-som-decay-rate-abalation-study-cifar100-SwinV2-S} were visualized to plot the training curves, which are shown in Figure~\ref{fig-SOM-cifar100-swinv2-train-loss-vs-iter-step312}. When plotting the function curves, downsampling and exponential weighted smoothing (seen \ref{appendix-sec1} for the specific code) were used to more intuitively compare the different convergence characteristics due to the different second-order moment estimate exponential decay rates. As seen in Figure~\ref{fig-SOM-cifar100-swinv2-train-loss-vs-iter-step312}, when the number of iterations was approximately less than $19,344$, the convergence rates of the four are similar. When the number of iterations was approximately greater than $19,344$, the $\overset{\frown}{\beta}_{2,t}$ of AdaX and our $\overset{\frown}{\beta}_{2,t}$ converged faster, indicating that $\overset{\frown}{\beta}_{2,t \to \infty} \to 1$ helps to accelerate convergence in the later stages of training.

\begin{figure}[H]
	\centering
	\subfloat[Training loss \emph{vs.} epoch]{\includegraphics[width=0.8\linewidth]{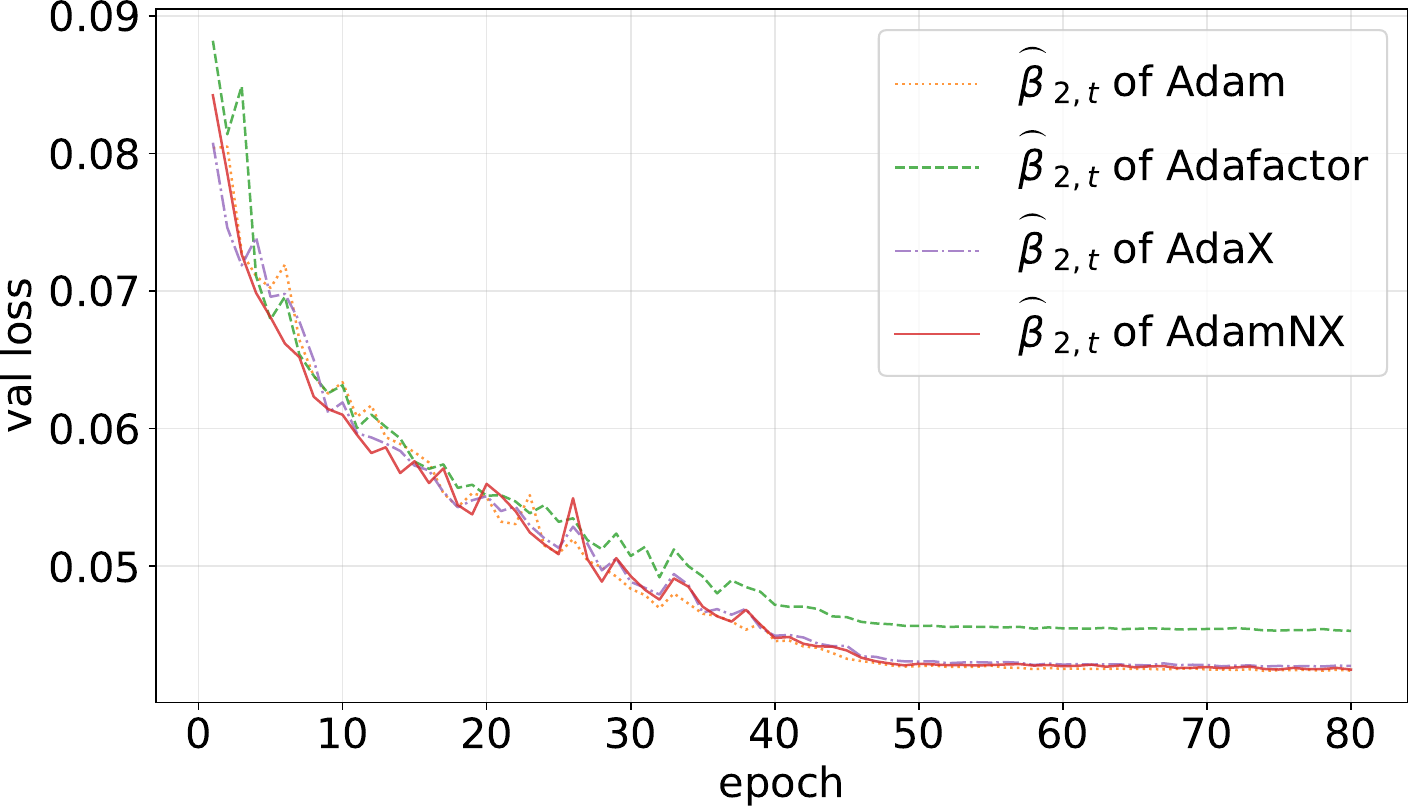}%
		\label{fig-SOM-voc2007-yolov7-tiny-val-loss-vs-epoch}}
	\\
	\subfloat[mAP@0.5 \emph{vs.} epoch]{\includegraphics[width=0.8\linewidth]{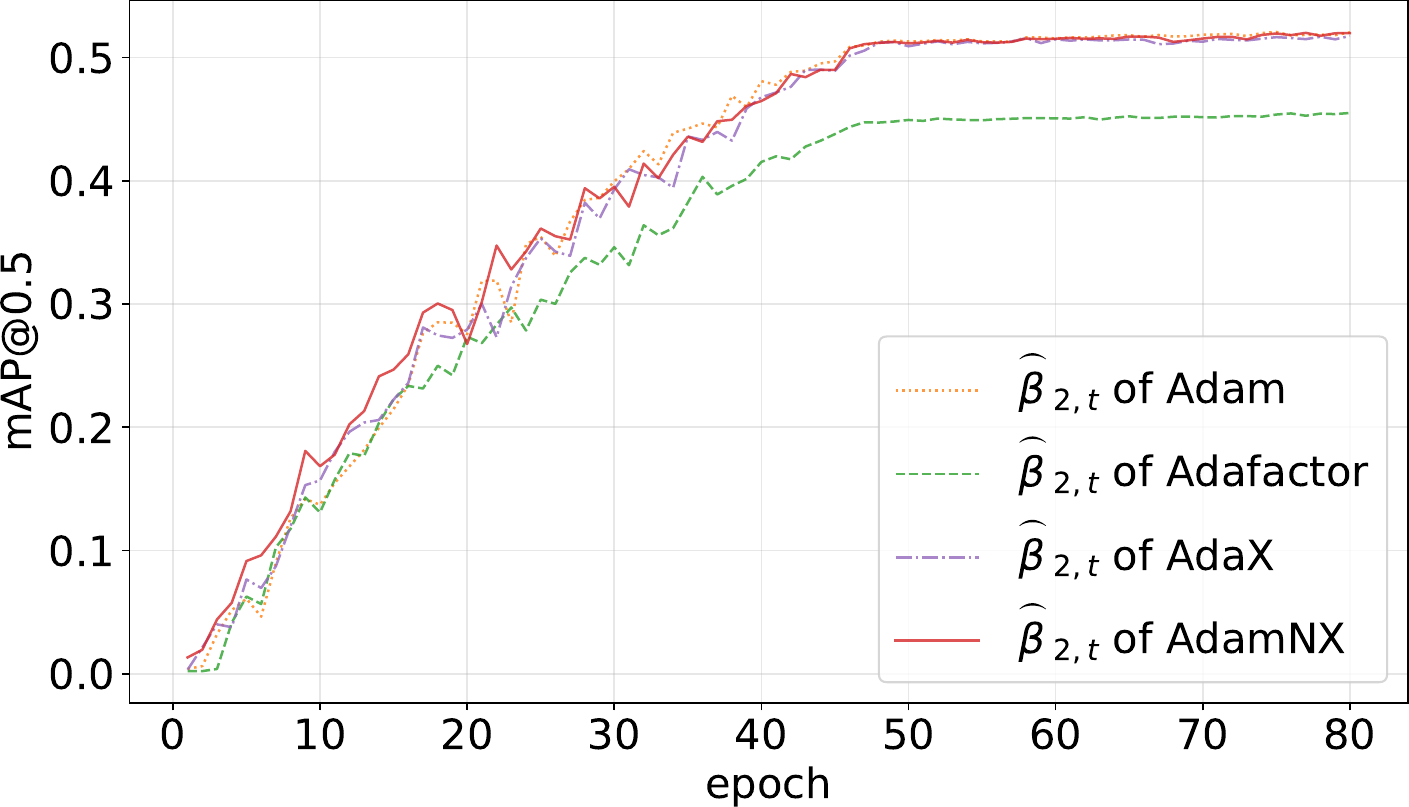}%
		\label{fig-SOM-voc2007-yolov7-tiny-mAP-0.5-vs-epoch}}
	\caption{Evaluation curve comparison}
	\label{fig-evaluation-curves-comparison}
\end{figure}

Comparison of object detection evaluation curves. The experimental logs from Table~\ref{tbl-som-decay-rate-abalation-study-VOC2007} were plotted as curves to further analyze the convergence characteristics of different optimization algorithms, and the results are shown in Figure~\ref{fig-evaluation-curves-comparison}. As seen in this figure, throughout the entire training process, the $\overset{\frown}{\beta}_{2,t}$ of Adam, the $\overset{\frown}{\beta}_{2,t}$ of AdaX, and the $\overset{\frown}{\beta}_{2,t}$ of this paper all converged faster and better than the $\overset{\frown}{\beta}_{2,t}$ of Adafactor, indicating the effectiveness of \enquote{for all $t \ge 1$, it holds that: $\overset{\frown}{\beta}_{2,t} \ge \overset{\frown}{\beta}_{1,t}$}.

\section{Comparison results and analysis of optimization algorithms of different type}\label{sec6}

This section compares AdamNX with momentum SGD~\cite{2013-Sutskever-SGDM}, RAdam~\cite{2020-Liu-RAdam}, Lion~\cite{2023-Chen-Lion}, and SophiaG~\cite{2024-Liu-Sophia}. The total number of training epochs was $80$ and the batch size was $64$. Each method uses the configuration reported here. Systematic tuning and multi-seed statistics were not used. Therefore, the comparison is limited to these configurations.

Table~\ref{tbl-cifar100-EfficientNetV2-L} shows the comparative experimental results on the benchmark model EfficientNetV2-L. For momentum SGD, the momentum was set to $0.9$, the peak learning rate was set to $0.1$, the minimum learning rate was set to $1 \times 10^{-3}$, $t_1$ was set to $37,488$, and the weight decay rate was set to $1 \times 10^{-3}$; for Lion, the peak learning rate was set to $1 \times 10^{-3}$, the minimum learning rate was set to $1 \times 10^{-5}$, $t_1$ was set to $37,488$, and the weight decay rate was set to $0.1$; for RAdam, SophiaG, and AdamNX, the peak learning rate was set to $0.01$, the minimum learning rate was set to $1 \times 10^{-4}$, $t_1$ was set to $37,488$, and the weight decay rate was set to $0.01$. As seen in Table~\ref{tbl-cifar100-EfficientNetV2-L}, our AdamNX achieved the lowest top-1 error rate of $35.66\%$ and the lowest top-5 error rate of $12.37\%$ on EfficientNetV2-L.

\begin{table}[H]
	\scriptsize
	\caption{Image classification results on CIFAR-100 for different types of optimization algorithms (benchmark model is EfficientNetV2-L)}
	\label{tbl-cifar100-EfficientNetV2-L}
	\begin{tblr}{
			width=\linewidth, 
			colspec={X[l] X[c] X[c]}, 	
			hline{1,Z}={1pt},                    
			hline{2}={0.5pt},                 	 
			columns={valign=m},                  
			rowsep=0.5pt                         
		}
		& Top-1 err. ($\%$) $\downarrow$ 
		& Top-5 err. ($\%$) $\downarrow$ \\
		momentum SGD~\cite{2013-Sutskever-SGDM}  & 39.03 & 16.19 \\
		RAdam~\cite{2020-Liu-RAdam}  & 36.12 & 12.66 \\
		Lion~\cite{2023-Chen-Lion}  & 41.33 & 17.03 \\
		SophiaG~\cite{2024-Liu-Sophia} & 41.01 & 15.42 \\
		AdamNX (Ours) & \textbf{35.66} & \textbf{12.37} \\
	\end{tblr}
\end{table}

Table~\ref{tbl-cifar100-ConvNeXt-B} shows the comparative experimental results with the baseline model being ConvNeXt-B. To adapt to the image resolution in CIFAR, ConvNeXt-B only retained the last three downsamplings, all convolutional kernel sizes were set to $(3, 3)$, and the padding was set to $(1, 1)$. For momentum SGD, the momentum was set to $0.9$, the peak learning rate was set to $0.01$, the minimum learning rate was set to $1 \times 10^{-4}$, $t_1$ was set to $37,488$, and the weight decay rate was set to $1 \times 10^{-3}$; for Lion, the peak learning rate was set to $1 \times 10^{-4}$, the minimum learning rate was set to $1 \times 10^{-6}$, $t_1$ was set to $37,488$, and the weight decay rate was set to $0.1$; for RAdam, SophiaG, and AdamNX, the peak learning rate was set to $1 \times 10^{-3}$, the minimum learning rate was set to $1 \times 10^{-5}$, $t_1$ was set to $37,488$, and the weight decay rate was set to $0.01$. As shown in Table~\ref{tbl-cifar100-EfficientNetV2-L}, our AdamNX achieved the lowest top-1 error rate of $32.87\%$ and the lowest top-5 error rate of $10.72\%$ on ConvNeXt-B.

\begin{table}[H]
	\scriptsize
	\caption{Image classification results on CIFAR-100 for different types of optimization algorithms (benchmark model is ConvNeXt-B)}
	\label{tbl-cifar100-ConvNeXt-B}
	\begin{tblr}{
			width=\linewidth, 
			colspec={X[l] X[c] X[c]}, 	
			hline{1,Z}={1pt},                    
			hline{2}={0.5pt},                 	 
			columns={valign=m},                  
			rowsep=0.5pt                         
		}
		& Top-1 err. ($\%$) $\downarrow$ 
		& Top-5 err. ($\%$) $\downarrow$ \\
		momentum SGD~\cite{2013-Sutskever-SGDM}  & 63.83 & 33.66 \\
		RAdam\cite{2020-Liu-RAdam}  & 35.45 & 11.95 \\
		Lion\cite{2023-Chen-Lion}  & 35.60 & 12.54 \\
		SophiaG\cite{2024-Liu-Sophia} & 37.39 & 13.48 \\
		AdamNX (Ours) & \textbf{32.87} & \textbf{10.72}  \\
	\end{tblr}
\end{table}

\section{Conclusion}\label{sec7}

This paper proposes AdamNX and the time-varying second-moment coefficient $\overset{\frown}{\beta}_{2, t} = \left(1 - \beta_2^{(1-\beta_2)(t-1)}\right) / \left(1 - \beta_2^{(1-\beta_2)t}\right)$. Under the assumptions used in our analysis, the coefficient gradually weakens update-scale correction and makes plateau-phase updates approach momentum-SGD-like behavior. The reported visual-task results show the tabulated values for AdamNX in the listed configurations and against the listed comparators. This paper does not establish convergence rates, regret bounds, general stability, flatness, statistical significance, engineering overhead, or cross-task generalization; the mathematical correctness of the relevant recurrence remains subject to author verification.

\section*{Acknowledgments}

This work was partly supported by the National Natural Science Foundation of China (Grant No. 62076117) and the Jiangxi Provincial Key Laboratory of Virtual Reality (Grant No. 2024SSY03151).

\appendix

\section{Downsampling and exponentially weighted smoothing code}\label{appendix-sec1}

\begin{lstlisting}[language=Python]
import pandas as pd
from matplotlib import pyplot as plt
from pathlib import Path

from plotting_config import *
from som_decay_rate_config import *

log_dir = Path("./excels")
T = 62480
step = max(1, T // 400) 
window = max(5, step * 0.02)
alpha = 2 / (window + 1)

plt.rcParams.update(plt_update)
fig, ax = plt.subplots(figsize=figsize)

for i, file in enumerate(log_dir.glob("*.xlsx")):
	df = pd.read_excel(file)

	ds = df.iloc[::step]   
	smooth = ds['loss'].ewm(alpha=alpha).mean() 
	file_stem = file.stem.split(" ")[-1]
	label = r"$\overset{\frown}{\beta}_{2,t}$" + " of " + file_stem   
	plt.plot(range(1, len(smooth)+1), smooth, linewidth=1.2, alpha=0.8, label=label,
					linestyle=linestyles[file_stem], color=colors[file_stem])


plt.xlabel("iteration $t$")

plt.ylabel('training loss (smoothed)')
plt.grid(alpha=0.3)
plt.legend()
plt.tight_layout()

save_path = f"pdfs/som-decay-rate-train-loss-vs-iter-step{step}.pdf"
fig.savefig(save_path, bbox_inches='tight', pad_inches=0)

plt.close()
\end{lstlisting}

\end{document}